\pdfoutput=1

\documentclass[11pt]{article}

\usepackage[]{acl}

\usepackage{authblk}

\usepackage{times}
\usepackage{latexsym}

\usepackage[T1]{fontenc}

\usepackage[utf8]{inputenc}

\usepackage{graphicx}
\usepackage{caption}

\usepackage{microtype}

\title{
    Building the Intent Landscape of Real-World Conversational Corpora with Extractive Question-Answering Transformers
}

\author[$\diamond$]{Jean-Philippe Corbeil\thanks{\hspace{0.5cm}jean-philippe.corbeil@polymtl.ca}}
\author[$\dagger$]{Mia Taige Li\thanks{\hspace{0.5cm}\{mia.li, hadi.abdi\_ghavidel\}@bell.ca}}
\author[$\dagger$]{Hadi Abdi Ghavidel}
\affil[$\dagger$]{Bell Canada}

\begin{document}
\maketitle
\begin{abstract}
    For companies with customer service, mapping intents inside their conversational data is crucial in building applications based on natural language understanding (NLU). Nevertheless, there is no established automated technique to gather the intents from noisy online chats or voice transcripts. Simple clustering approaches are not suited to intent-sparse dialogues. To solve this intent-landscape task, we propose an unsupervised pipeline that extracts the intents and the taxonomy of intents from real-world dialogues. Our pipeline mines intent-span candidates with an extractive Question-Answering Electra model and leverages sentence embeddings to apply a low-level density clustering followed by a top-level hierarchical clustering. Our results demonstrate the generalization ability of an ELECTRA large model fine-tuned on the SQuAD2 dataset to understand dialogues. With the right prompting question, this model achieves a rate of linguistic validation on intent spans beyond 85\%. We furthermore reconstructed the intent schemes of five domains from the MultiDoGo dataset with an average recall of 94.3\%.
\end{abstract}

\section{Introduction}

With the rise of pre-trained NLU models in the last few years, call centers now have robust tools in their reach to optimize their customer operations. Two main applications are chatbots and head-intent detection, of which both need to define a taxonomy of intents relevant to the business. In most cases, it is an overwhelming task for companies, despite having a readily available amount of transcripts and chats.

Current approaches in the field of intent discovery \cite{popov-etal-2019-unsupervised,vedula2019towards,vedula2020automatic,chatterjee2020intent,zhang2021textoir,hang2021clustering} tend to focus on datasets with intent-dense utterances --- e.g. ATIS \cite{hemphill1990atis} or SNIPS \cite{coucke2018snips}. However, this type of dataset is rarely available when starting the design of NLU applications. Usually, companies have dialogues between an agent and a customer, of which most utterances do not have a clear intent prompt. We characterize these real-world datasets as intent-sparse. We propose the \textit{intent landscape} as the task to extract the intents and the taxonomy of intents from intent-sparse dialogue datasets. We consider this task as a generalization of the \textit{intent discovery} task, which focuses on labelled and unlabelled utterance corpora to find known and unknown intents.

How can we automatically extract intents from a noisy intent-sparse dialogue corpus? We propose a pipeline to extract intent spans from conversational data into a data-driven intent hierarchy to solve this.

Our contributions are five folds:

\begin{enumerate}
    \item We defined the intent landscape task.
    \item We designed a pipeline that extracts intent spans from real-world dialogues and maps the relevant intents into a hierarchy of clusters.
    \item We proposed a strategy to estimate the count of each cluster based on semantic similarity.
    \item We show that the ELECTRA large \cite{clark2020electra} fine-tuned on the SQuAD2 Question-Answering dataset \cite{rajpurkar2018know} has the ability to extract relevant intent spans from conversations.
    \item We observed flaws in the original intent schemes of the MultiDoGo dataset \cite{peskov2019multi}, and we discovered a few new intent clusters in each domain.
\end{enumerate}

We first review the previous works. We present our methodology afterwards: the dataset, our pipeline and the experiments. We then report each of our results and discuss our main findings. We conclude with a few closing remarks.

\section{Related Work}
When looking for intents in text, one usually needs to assess whether open intents exist, i.e. a predefined taxonomy can confine all intents in the scope of the current dataset. With the assumption that there exist no open intents, intent detection becomes a supervised classification task \cite{vedula2019towards}. Some works on this front focus on finding and analyzing intents from social media such as online forums and microblogs \cite{agarwal2017tumblr,gupta2014purchase, wang2015twitter, chen-etal-2013-identifying}; while others look at intent detection as part of Spoken Language Understanding, and study it jointly with slot filling. A typical architecture used in this field is attention-based RNN \cite{goo-etal-2018-slot, mesnil2015slot, zhang2016slot}, and in recent years many explored the benefits of adversarial learning \cite{kim-etal-2017-adversarial, liu2017slot, yu2018adversarial}. 

On the other hand, many previous works focus on capturing unknown intents in given data. According to \citet{zhang2021textoir}, there are two tasks on this end: open intent detection and open intent discovery\footnote{The terms "intent discovery" and "intent mining" have been used interchangeably in the literature, both referring to the task of identifying types of unknown intents in text. We will use "intent discovery" in our paper.}. Open intent detection is an n+1-class classification with n known intent classes and one open intent class. Many works in this area use a threshold-based method for such decision \cite{hendrycks2016misclassified,shu2017doc,liang2017principled}, but there are also attempts to use geometrical features to alleviate the problem of relying on the presence of unknown intents in the train set \cite{lin-xu-2019-deep, zhang_xu_lin_2021}. However, the downside of this task is that there is no way to differentiate one open intent from another, as they are all placed into one umbrella category. 

Unlike previously-mentioned tasks, open intent discovery does not require a predefined intent taxonomy. Researchers in this field leverage both unsupervised and supervised learning and domain-expert knowledge. \citet{vedula2019towards} develops a pipeline incorporating attention-based LSTM and CRF, trains on data collected from Stack Exchange, to discover unknown intents by sequentially tagging the action-object pairs determined by the CRF; \citet{cai2017cnn} uses unsupervised clustering and domain knowledge on online medical forums to map out an intent taxonomy for classifying medical intents online. The limitation of these two works is that both rely on data in the form of short, intent-dense prompts. In reality, open intent discovery often needs to be done on noisy, intent-sparse data such as conversation dialogues. There are various approaches in the literature to tackle this problem. \citet{popov-etal-2019-unsupervised} takes intent discovery as a topic modelling task, which can process input data as a whole, therefore avoiding the need for intent-dense prompts. However, \citet{hang2021clustering} points out that this approach relies on additional linguistic features to help the performance, as conversational dialogues tend to be shorter in length and contain less latent semantic information when compared to documents and paragraphs used in topic modelling. \citet{chatterjee2020intent} uses a domain-agnostic pre-trained Dialog Act Classifier, where any utterances being tagged as QUESTION or INFORMATION by the classifier would be considered as potential intent candidates and get passed on into downstream clustering. Nonetheless, the assumption that we capture all intents within utterances tagged as one of these two classes limits their approach. \citet{vedula2020automatic} proposes a 3-stage system that can be trained on known intents in a known domain to leverage knowledge transfer and discover unknown intents in other domains, with the assumption that the chosen known intents and their relation with the domain are an appropriate representation of the unknown. 

Our work presents a cleaner way to extract unknown intents from conversation dialogues with fewer assumptions about their characteristics. We focus only on the syntactical integrity of intent phrases, and we do not assume anything about their semantics. Moreover, we also map the intents into relevant categories at many levels of granularity.

\section{Methodology}

\subsection{Dataset}

We need a conversational dataset to evaluate our pipeline with the following characteristics: covering a few domains, annotated intents, having two channels (e.g. agent and customer) and textual data about a real-life customer service interaction with multiple turns. In the literature, there is mainly three types of high-quality conversational datasets that are available for research purposes: intent-dense utterance datasets \cite{coucke2018snips,hemphill1990atis}, task-oriented dialogue datasets \cite{budzianowski2018multiwoz,peskov2019multi,eric2019multiwoz,zang2020multiwoz,rastogi2020towards,chen2021action} and open-dialogue datasets \cite{li2017dailydialog,rashkin2018towards,zhang2018personalizing}. We discard the open-dialogue datasets since we need labelled intents in the dataset. To work with actual conversations, we cannot rely on the intent-dense utterance dataset, which already considers a selection of certain types of utterances. Therefore, we selected the task-oriented dialogue datasets, which would closely match our criteria. Despite being cleaner than real-world conversations, the structure of these conversations will still emphasize intents from a customer (see Table \ref{tab:dialog_ex}). To augment the sparsity of intents and have real customer intentions, we do not take into account any intent that are general conversational markers (e.g. \textit{opening greeting}, \textit{closing greeting} and \textit{confirmation}) or system markers (e.g. \textit{out-of-domain}), and not head intentions. By doing this, we focus the evaluation of our pipeline on real domain-specific intents.

We chose the MultiDoGo dataset containing six different domains and multi-turn conversations. Its name stands for "\textbf{Multi}-\textbf{Do}main \textbf{Go}al-oriented dialogues" and it is a dataset compiled by the AWS Labs \cite{peskov2019multi}. It contains dialogues from 6 different domains: airline, media, insurance, finance, software, and fast food. We have three different sets of data: annotated at the turn level, annotated at the sentence level and unannotated. The first two were labelled only on the customer channel and are split into three sets: \textit{train}, \textit{dev} and \textit{test}. We leverage the test sets of this annotated data to extract the intent schemes for each domain. We discarded 7 conversational markers or technical markers that are present in all the intent sets: \textit{openinggreeting}, \textit{closinggreeting}, \textit{confirmation}, \textit{rejection}, \textit{contentonly}, \textit{thankyou} and \textit{outofdomain}. For our experiments, we are using the \textit{unannotated} dataset since it contains the full dialogues (customer and agent channels).

\begin{table}[ht!]
\caption{Example of dialogue formatted from the MultiDoGo \textit{media} dataset.}
\fbox{
    \small
    \begin{tabular}{ll}
        \textbf{customer:} & hello\\
        \textbf{agent:} & Hello there! Welcome to Inflamites\\
         & Cable/Media service, how may I help \\
         & you today?\\
        \textbf{customer:} & i want to purchase new cable service\\
        \textbf{agent:} & ... \\
    \end{tabular}
}
\label{tab:dialog_ex}
\end{table}

\subsection{Pipeline}

The pipeline is illustrated in Figure \ref{fig:engine_pipeline}. It is composed of five steps:

\begin{figure*}[ht!]
    \centering
    \includegraphics[width=\linewidth]{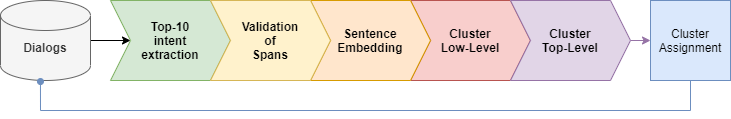}
    \caption{Diagram of the Intent-Landscape Pipeline.}
    \label{fig:engine_pipeline}
\end{figure*}

\begin{table*}[ht!]
    \caption{Information of Top Question-Answering Models fine-tuned on SQuAD2 selected on the HuggingFaceHUB.}
    \centering
    \begin{tabular}{|c|c|c|c|c|}
        \hline
        Model Size & Creator & Name & F1 & EM \\
        \hline
        small & {\small mrm8488} & {\small\verb|electra-small-finetuned-squadv2|} & 73.4 & 69.7 \\
        \hline
        base & {\small PremalMatalia}  & {\small\verb|electra-base-best-squad2|} & 83.2 & 79.3 \\
        \hline
        large & {\small ahotrod} & {\small\verb|electra_large_discriminator_squad2_512|} & 90.0 & 87.0 \\
        \hline
    \end{tabular}
    \label{tab:qa_model_performance}
\end{table*}

\begin{enumerate}
    \item \textbf{Intent Extraction:} Extract $N$ intent-like candidates based on the span extraction done with an extractive Question-Answering model.
    \item \textbf{Validation of Spans:} Check that the span corresponds to a phrase expressing an intent. We impose four main criteria: all candidates answer the question (i.e. no unanswerable token), Part-of-Speech contains \textit{action-object} phrase, the sentence has an appropriate form (i.e. length and clean formatting), and span comes from a customer utterance only.
    \item \textbf{Sentence Embedding:} Encode the spans into sentence embeddings.
    \item \textbf{Clustering Low-Level:} Apply a density-based clustering to find the meaningful clusters from densely packed regions. We assume that frequent fine-grained intent spans are relevant to consider as low-level intents.
    \item \textbf{Clustering Top-Level:} Apply a hierarchical clustering on the low-level cluster-center embeddings to establish the structure of the fine-grained intent spans into high-level intents.
\end{enumerate}

\subsubsection{Extractive Question-Answering Model}

The extractive Question-Answering (QA) task relies on two inputs: a question and a context paragraph. Both are provided to the transformer encoder with the SEP token as a delimiter. The goal is to extract a span of text --- start index to end index inside the token sequence --- from the context paragraph that answers the question. It is common to use a ranking strategy to consider the top $K$ potential answers. In our work, we leverage the pipeline implementation from HuggingFace to generate our candidates \cite{wolf-etal-2020-transformers}. We can usually answer the question with a span of text from the paragraph. However, the SQuAD2 dataset \cite{rajpurkar2018know} was built with unanswerable questions as well. The authors used the empty string (\textit{""}) as a span answer for a question that we cannot answer with the context paragraph. We refer in this paper to the unanswerable situation as "impossible" as in "impossible to answer". We use this in our validation step \textit{Ignore any Impossible} (see Section \ref{section:QA}).

\subsubsection{Dialogue Pre-processing}

To form the context paragraph, we concatenate the utterances at the conversation level according to their turn order. We also append the channel name at the beginning of each utterance with a column ":" in-between, and we append a line return at the end. We hypothesize that this format would help the question-answering transformer model leverage similar dialogue patterns seen during its pre-training. The dialogue string takes the appearance of the example in Table \ref{tab:dialog_ex}.

\subsubsection{Sentence Embedding}

We encode the candidate spans into sentence embeddings using the \textit{sentence-transformer} bi-encoder approach \cite{reimers-2019-sentence-bert}. It uses an encoding transformer (e.g. BERT) with a pooling layer on all contextual embeddings, trained in a Siamese fashion under the cosine similarity.

\subsubsection{Low-level Density Clustering}

Since we do not want a fixed amount of clusters and we want to focus on dense regions of the embedding space, we used the HDBSCAN algorithm \cite{mcinnes2017hdbscan} to determine the low-level clusters based on the cosine distance, given by the formula: $1\ -\ cosine\ similarity$. HDBSCAN has the advantage of capturing the hierarchical structure of the space to extract density clusters to determine the distance threshold of each cluster, which is a limitation of the DBSCAN algorithm \cite{ester1996density}. It is also faster than OPTICS \cite{ankerst1999optics}, which we experimented with and found closely match outputs in our pipeline. These density-clustering techniques are not relying on a fixed amount of clusters, but they filter out noise by focusing on clusters with more than \textit{min\_cluster\_size} examples. We use 2 by default in our experiment, but larger domains (airline and media) were less noisy with slightly higher numbers (4 and 3, respectively).

\subsubsection{Top-Level Hierarchical Clustering}

To build the hierarchy of intents on low-level cluster centers, we rely on hierarchical clustering \cite{ward1963hierarchical} by average link on cosine distance. We compute the low-level cluster centers using the average of all its members. The main hyperparameter is the \textit{distance\_threshold}, which sets the cutting point to form the top-level clusters. We manually tune it between $0.2$ and $0.5$, which translates roughly in cosine distance as many small clusters and few large clusters, respectively. We select the hyperparameter that gives the most relevant clusters visually on the TSNE 2D plot (e.g. each blob should have the same colour), and we validate that it keeps the clusters homogeneous semantically by inspection (e.g. all order intents should be a top-level cluster \textit{order}).

\subsubsection{Manual Mapping After Top-Level}

From the top-level clustering, we observe imperfections in the hierarchy because of the variability of the distance of discrimination between clusters --- i.e. some pairs of clusters tend to be closer than others from a semantic similarity perspective. Our experiments include a final manual step of cleaning the automated outputs from the pipeline based on semantic meaning, which includes merging similar clusters and grouping one cluster contained in another. We also included that step to match the intents in the test sets for evaluation purposes. We let the furthermore automation and refinement of this step as future work.

\subsubsection{Count Estimation Strategy}

Most of the time, we still have many dialogues that we cannot attach to an intent after the clustering. We observed in our experiment that it is usually between 10\%-25\% of the dialogues. To complete the estimation of intent volumes, we apply a completion strategy by passing through the candidate spans of these dialogues, and we look at the similarity between these and the low-level cluster centers. We consider the low-level cluster centers since they are specific and give a more precise similarity. We take the minimum similarity scores across the cluster centers, and if it is below some threshold, we have found a cluster to assign to this dialogue. We named that threshold the \textit{force\_cluster\_threshold}. We can tune this threshold given the amount of noise we can tolerate. In our experiments, we aim to keep accurate and homogeneous clusters. Thus, we keep that threshold around $0.2$ and $0.3$.

\subsection{Experiments}

\subsubsection{Question-Answering Validation Experiment}
\label{section:QA}

First, we experiment with the QA models to understand the boundaries of its application in the intent-landscape pipeline. We change two parameters --- model size and question prompting --- and measure their impact on the intent linguistic validation rates in the validation step of the pipeline. For the model sizes, we considered all three sizes of the ELECTRA model \cite{clark2020electra}: small, base and large. We selected models fine-tuned on the SQuAD2 dataset \cite{rajpurkar2018know} based on the highest F1 score from the HuggingFaceHUB\footnote{https://huggingface.co/models} \cite{wolf-etal-2020-transformers} in Table \ref{tab:qa_model_performance}.

Nonetheless, we still need to articulate well a question to query the intent-like spans from the customer. Thus, we select three different prompting questions:

\begin{itemize}
    \item \textbf{Q1}: What is the main reason of the call mentionned by the customer?
    \item \textbf{Q2}: What can the agent help the customer with?
    \item \textbf{Q3}: What is the customer's first intent?
\end{itemize}

With \textbf{Q1}, we emphasize what the customer mentions during the call, and we specify that we are looking for the "main reason". On the other hand, we formulate \textbf{Q2} to look for what the agent can do for the customer using the verb "help". At last, the last prompt \textbf{Q3} ignores the call aspect and only asks for the customer's "first intent" in technical terms.

\begin{figure}[ht!]
    \centering
    \includegraphics[width=\linewidth]{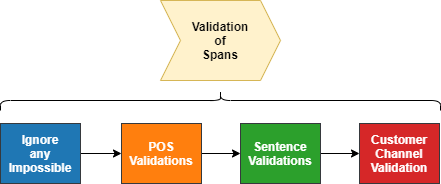}
    \caption{Validations applied to span extracted by the extractive Question-Answering model.}
    \label{fig:valid_steps}
\end{figure}

 We quantify the quality of the extraction and the generalization of the QA model to the dialogue domain based on the \emph{Validation of Spans} in Figure \ref{fig:engine_pipeline}. In Figure \ref{fig:valid_steps}, we break down that step into four components:
 
 \begin{itemize}
     \item \emph{Ignore any Impossible}: Remove the dialogue if any of its candidate spans is the impossible marker in the top $K$ extracted by the QA model.
     \item \emph{POS validations}: Validate the candidate span based on what we expect from an intent span. We assume that there will be the presence of both an \textit{action} and an \textit{object}. We relate these to a VERB and a NOUN Parts-of-Speech, respectively. We leveraged industrial-grade English POS tagger from Spacy \cite{spacy2}. For instance, we expect the intent text spans to be extracted similarly to "I [want]$_{VERB}$ black [smartphone]$_{NOUN}$". Moreover, the pronoun "I" could be dropped in this example, which would still be considered valid.
     \item \emph{Sentence validations}: We check that the span does not contain any dialogue format artefacts like channel prefixes (\textit{"customer: "} and \textit{"agent: "}) as well as a line return. Afterwards, we remove any span below 2 tokens or beyond 12 tokens based on whitespace tokenization. We assume that only one word cannot be an intent, and a span above 12 tokens means a lack of precision from the QA model.
     \item \emph{Customer Channel Validation}: Remove the span candidate if it is not from the customer channel of the dialogue. 
 \end{itemize}

 We count all the remaining dialogues containing an intent-candidate span after applying all validations sequentially. Then, we calculate the absolute percentages by dividing by the initial count of dialogues for all validation steps. We consider a combination of a model and a prompt question better if it finds the most dialogues with at least one valid intent span.

\subsubsection{Intent Scheme Recovery Experiment}

In this experiment, we apply the pipeline in Figure \ref{fig:engine_pipeline} on five of the six domains in MultiDoGo. We consider that the ground-truth intents are associated with our top-level intents extracted by the pipeline. To validate the pipeline experimentally, we need to find these associations. Therefore, we manually assess the mapping from the resulting top-cluster spans to the known domain-specific intents. We recover the domain intents from the labelled test set data and remove the conversational and system markers. We map the extracted top-level intent spans (e.g. \textit{i need to my  seat assignment}) based on how close it is from an actual domain intent (e.g. \textit{getseatinfo}). If it means the same or the span can be in its scope without any other match, we associate them. Otherwise, we assigned the \textit{OTHER} intent, which is our marker for newly discovered intent. Two authors participated in this effort, and a review session set the final associations. Finally, we compute the recall of intents if at least one top-level span matched that intent.

\subsubsection{Cluster-Quality Experiment}

 To demonstrate the quality of the extracted clusters, we report the classification metrics \textbf{P}recision, \textbf{R}ecall and \textbf{F1}-score on the annotated test sets. We consider only the intents with support above 10 --- to ensure significant results. We use the semantic similarity based on cosine as a zero-shot classification applied on the set of valid span candidates $\vec{s}$ after both the QA  and the validation step (see Figure \ref{fig:engine_pipeline}). 
 
 We select the estimate intent $\hat{y}$ by picking the most similar low-level cluster center $\vec{c_i}$ (see equation \ref{eq:1}). We assign "unlabeled" to any example with all similarity scores below $0.4$. Then, we trace the top-level cluster from the low-level one using the taxonomy. To get our final metrics, we use our manual mapping to compute the classification scores with the ground truth (see Appendix \ref{sec:appendix}).

\begin{equation}
    \label{eq:1}
    \hat{y} = \mathrm{argmax}_i\left(\frac{\vec{s} \cdot \vec{c_i}}{||\vec{s}|| ||\vec{c_i}||}\right)
\end{equation}

\section{Results}

In this section, we present the results of three experiments: question-answering extraction by linguistic validations, intent scheme recovery by our intent-landscape pipeline, and validation of the quality of intent-span clusters through zero-shot similarity classification.

\subsection{Question-Answering Validation Experiment}

\begin{figure*}[ht!]
    \centering
    \includegraphics[width=\linewidth]{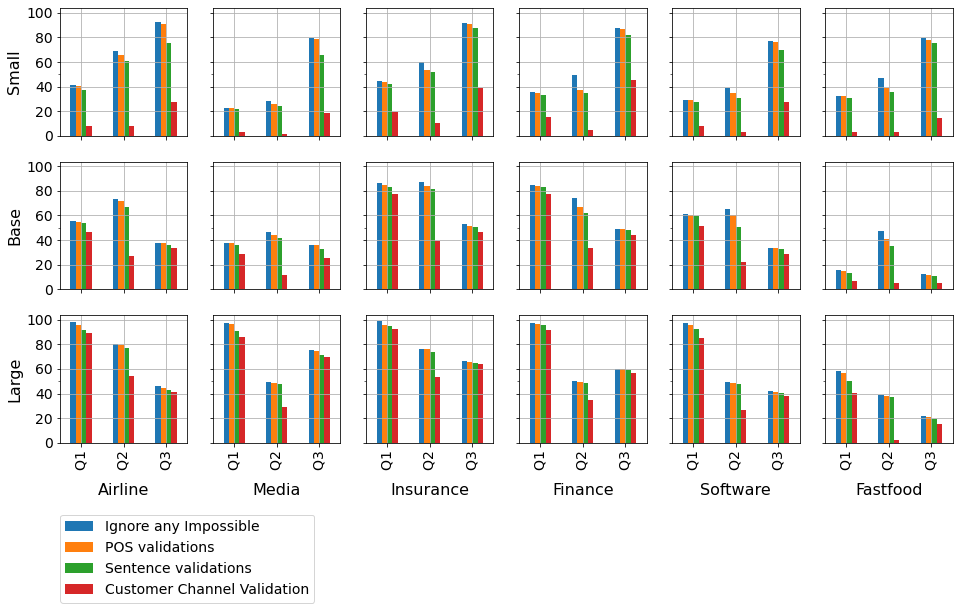}
    \caption{Validations done on the 10 intent-like span candidates for all MultiDoGo dialogues extracted with the ELECTRA models fine-tuned on SQuAD2 (small, base and large). The absolute percentages are computed based on the presence of any valid candidate for each conversation remaining after each validation divided by the initial amount of conversations.}
    \label{fig:valid_analysis}
\end{figure*}

We observe drastic changes in the results in Figure \ref{fig:valid_analysis} for the various combinations of model sizes, the prompting questions and domains. The large model achieves the best combination with the first prompt \textbf{Q1}, which is formulated with general terms like "main reason of the call" and asks to focus on the "customer" only. This combination performs a final validation rate above 85\%, except in the fast-food domain. On the other hand, the small models can achieve a considerable validation rate with the last prompt \textbf{Q3} right before the stage named \textit{Customer Channel Validation}. However, it fails at this last validation step. Therefore, we hypothesize that the small model size does not have a good representation of a dialogue structure in a text, which is more abstract and should come in the latest layers. With the wrong prompting, its performances are also less reliable. Our results show that the base model can perform considerably in the insurance and finance domains. In the other domains, this model size seems to struggle by relying a lot on the Question-Answering unanswerable span (i.e. ""), which reduces the number of dialogues from the first validation step \textit{Ignore any Impossible}. In general, \textbf{Q2} prompt performed the worst at the last stage of our validation process, i.e. \textit{Customer Channel Validation}. We hypothesize that mentioning the agent and the customer inside the question confuses the model.

Our results highlight that the fast-food domain is more challenging. When observing the intents and the dialogues, we noticed that the agents and the customers use particular terms around the meals. On top of this, we further noted similar sentence structures about ordering food. Thus, we argue that these two aspects make it difficult for the transfer learning of the QA models to generalize on that domain. We speculate that conversational design grounded in the best practices seems only to contain one order intent and the named entity extraction on the meal names.

In section \ref{sec:intent-recovery-exp}, we base our experiments on the best combination formed by the large model with the first question \textbf{Q1}. We also ignore the fast-food domain for the reasons mentioned above.

\subsection{Intent Scheme Recovery Experiment}
\label{sec:intent-recovery-exp}

In Table \ref{tab:intent_map}, we summarized the results regarding our findings on the intents of all five domains using our pipeline. Overall, we observed a recall of 94.3\% on average, which indicates that we found the vast majority of intents. Our pipeline only missed one intent in media and one in software. We also discovered between 3 and 6 new intents. We shared in the Appendix \ref{sec:appendix} our mapping, the TSNE 2D plots of the clusters and the clustering hyperparameters.

\begin{table}[ht!]
    \caption{Summary of results with the mappings based on the top clusters, compared with intent schemes (\textit{Total} column ignoring intents that are conversation markers or \textit{outofdomain}). The \textit{Total}, \textit{Found} and \textit{New} are numbers of intents. The recall of found intents by our pipeline in the original intent scheme is the percent of \textit{found}/\textit{total}.}
    \centering
    \begin{tabular}{|c|c|c|c|c|}
        \hline
        \textbf{Domain} & \textbf{Total} & \textbf{Found} & \textbf{New} & \textbf{Recall} \\
        \hline
        airline & 4 & 4 & 4 & 100\% \\
         \hline
        media & 6 & 5 & 3 & 83\% \\
        \hline
        insurance & 3 & 3 & 6 & 100\% \\
        \hline
        finance & 10 & 10 & 5 & 100\% \\
        \hline
        software & 8 & 7 & 5 & 88\% \\
        \hline
    \end{tabular}
\label{tab:intent_map}
\end{table}

\subsection{Cluster-Quality Experiment}

\begin{table}[ht!]
\caption{Airline report on selected intents (support above 10).}
\label{table:airline}
\centering
\begin{tabular}{|c|c|c|c|}
\hline
Intent & P & R & F1 \\
\hline
\small{changeseatassignment} & 0.92 & 0.33 & 0.48\\
\hline
\small{getboardingpass} & 0.96 & 1.0 & 0.98\\
\hline
\small{bookflight} & 0.89 & 0.96 & 0.92\\
\hline
\small{getseatinfo} & 0.15 & 0.93 & 0.26\\
\hline
\end{tabular}
\end{table}

\begin{table}[ht!]
\caption{Media report on selected intents (support above 10).}
\label{table:media}
\centering
\begin{tabular}{|c|c|c|c|}
\hline
Intent & P & R & F1 \\
\hline
\small{startserviceintent} & 0.8 & 0.68 & 0.73\\
\hline
\small{viewbillsintent} & 0.95 & 1.0 & 0.97\\
\hline
\end{tabular}
\end{table}

\begin{table}[ht!]
\caption{Insurance report on selected intents (support above 10).}
    \label{table:insurance}
    \centering
    \begin{tabular}{|c|c|c|c|}
        \hline
        Intent & P & R & F1 \\
        \hline
        \small{checkclaimstatus} & 0.91 & 0.98 & 0.95\\
        \hline
        \small{getproofofinsurance} & 0.98 & 1.0 & 0.99\\
        \hline
        \small{reportbrokenphone} & 0.76 & 1.0 & 0.86\\
        \hline
    \end{tabular}
\end{table}

\begin{table}[ht!]
\caption{Finance report on selected intents (support above 10).}
\label{table:finance}
\centering
\begin{tabular}{|c|c|c|c|}
\hline
Intent & P & R & F1 \\
\hline
\small{reportlostcard} & 0.82 & 1.0 & 0.9\\
\hline
\small{updateaddress} & 0.9 & 1.0 & 0.95\\
\hline
\small{checkbalance} & 0.97 & 0.99 & 0.98\\
\hline
\small{transfermoney} & 0.91 & 0.97 & 0.94\\
\hline
\small{disputecharge} & 0.67 & 0.98 & 0.8\\
\hline
\end{tabular}
\end{table}

\begin{table}[ht!]
\caption{Software report on selected intents (support above 10).}
\label{table:software}
\centering
\begin{tabular}{|c|c|c|c|}
\hline
Intent & P & R & F1 \\
\hline
\small{reportbrokensoftware} & 0.64 & 0.97 & 0.77\\
\hline
\small{softwareupdate} & 0.9 & 0.49 & 0.63\\
\hline
\small{expensereport} & 0.62 & 0.96 & 0.75\\
\hline
\small{startorder} & 0.61 & 0.95 & 0.74\\
\hline
\small{checkserverstatus} & 0.88 & 0.91 & 0.9\\
\hline
\end{tabular}
\end{table}

 We displayed the classification results for all domains in Tables \ref{table:airline}, \ref{table:media}, \ref{table:insurance}, \ref{table:finance} and \ref{table:software}. We note that most of the \textbf{P}recisions, \textbf{R}ecalls and \textbf{F1}-scores are above $0.9$, which is considerable for a zero-shot setting and indicates a high-quality clustering. There are a few lower values that are related to the incompatibility between our pipeline and the flat taxonomy design of the MultiDoGo dataset.

In the following paragraphs, we use the centered dot "$\cdot$" to indicate that the intents on both sides have a relatively high similarity score compared to other intents in the domain. Two pairs of intents that particularly caught our attention are \textit{changeseatassignment} $\cdot$ \textit{getseatinfo} from the airline domain, and \textit{startserviceintent} $\cdot$ \textit{getinformationintent} from the media domain. In the former case, both intents frequently contain the phrase "seat assignment", making the two semantically similar. In the latter case, both intents are syntactically similar to the patterns "I want …" and "I'd like to …". Due to their semantic or syntactical similarities, the distances between the clusters of these pairs of intents are much closer than compared to those of other intents in the same domain. On top of the flat taxonomy design of the MultiDoGo dataset, it is more difficult for the pipeline to distinguish these pairs of similar intents apart when the other intents are much further away from each other. As a result of this problem, we discovered that utterances such as "I would like to buy 10 Gb plan", which belongs to \textit{startserviceintent}, are labelled as \textit{getinformationintent}.

This aforementioned issue can be easily solved by constructing a two-stage taxonomy, in which the first level maximizes the dissimilarity between intents at a high level, and the second level targets more fine-grained distinctions. For example, in the software domain the intents can be naturally defined into three broader groups based on their similarity: \textit{softwareupdate} $\cdot$ \textit{reportbrokensoftware} $\cdot$ \textit{checkserverstatus}, \textit{expensereport} $\cdot$ \textit{getpromotions}, and \textit{startorder} $\cdot$ \textit{changeorder} $\cdot$ \textit{stoporder}.


\section{Conclusion}

In conclusion, we proposed a pipeline to solve our intent-landscape task by extracting the intents and an intent taxonomy from a corpus composed of real-life customer-service dialogues. First, we experimented with the first two stages that extract and validate intent-like spans from dialogues. We showed the generalization ability of the Question-Answering Electra large model to a dialogue context, with a linguistic validation of the extracted intent spans above 85\%. We recovered the intent schemes from five domains with our pipeline at an average recall of 94.3\%. At last, we suggested that a two-level taxonomy could alleviate flaws in airline, media, and software schemes. We have seen two limitations with particular domains like fast food and the manual mapping step after our top-level clusters. We let these two as future works. Reasonable approaches to tackling these two issues could be leveraging domain adaptation and upgrading semantic similarity with natural language inference.

\section*{Acknowledgements}

We thank Bell Canada (BCE inc.) as well as everyone involved: Nassim Guerroumi, Ryan Levman, Stephanie Maccio, Jeff Kurys, Michel Richer, and Alan Khalil.

\bibliography{main}
\bibliographystyle{acl_natbib}

\appendix

\newpage
\section{Appendix}
\label{sec:appendix}

\begin{table}[ht!]
\caption{Manual mapping for \textit{airline} from top clusters to intents with estimated volumes on test set.}
\label{table:airline_map}
\centering
\begin{tabular}{|p{2.5cm}|c|c|}
\hline
Top Cluster & Intent & Volume \\
\hline
\small{can you pls send me my boarding pass} & \small{getboardingpass} & 101\\
\hline
\small{i need to check my seat assignment} & \small{getseatinfo} & 91\\
\hline
\small{i want book a flight ticket} & \small{bookflight} & 90\\
\hline
\small{i need to check my seat} & \small{getseatinfo} & 49\\
\hline
\small{i already book a ticket but i change this seat} & \small{changeseatassignment} & 23\\
\hline
\small{i need to change seat} & \small{changeseatassignment} & 17\\
\hline
\small{i need your help} & \small{OTHER} & 13\\
\hline
\small{i need to my  seat assignment} & \small{getseatinfo} & 13\\
\hline
\small{boarding pass to be sent to my email address} & \small{getboardingpass} & 11\\
\hline
\small{i wanna change my seat assignment} & \small{changeseatassignment} & 10\\
\hline
\small{could you tell me my seat arrangement} & \small{getseatinfo} & 9\\
\hline
\small{i need middle seat} & \small{changeseatassignment} & 8\\
\hline
\small{i need ticket for chennai} & \small{bookflight} & 5\\
\hline
\small{i need a boarding pas} & \small{getboardingpass} & 5\\
\hline
\small{i want to know saet no} & \small{getseatinfo} & 3\\
\hline
\small{i have an upcoming flight} & \small{OTHER} & 3\\
\hline
\small{i already book a ticket but i want to chnage it} & \small{OTHER} & 3\\
\hline
\small{i want to book one way ticket} & \small{bookflight} & 2\\
\hline
\small{i need flight under \$300} & \small{bookflight} & 1\\
\hline
\small{i want to changing the window seat} & \small{changeseatassignment} & 1\\
\hline
\small{i need a plan in ticked booking} & \small{OTHER} & 1\\
\hline
\small{i wand a plain  ticket} & \small{bookflight} & 1\\
\hline
\small{i am need a flight from newyork to texas} & \small{bookflight} & 1\\
\hline
\end{tabular}
\end{table}

\begin{table}[ht!]
\caption{Manual mapping for \textit{media} from top clusters to intents with estimated volumes on test set.}
\label{table:media_map}
\centering
\begin{tabular}{|p{2.5cm}|c|c|}
\hline
Top Cluster & Intent & Volume \\
\hline
\small{i want new cable service connection} & \small{startserviceintent} & 192\\
\hline
\small{i like to purchase a new internet connection service} & \small{startsertviceintent} & 125\\
\hline
\small{i want internet plan} & \small{startserviceintent} & 59\\
\hline
\small{i would like to sign up new service} & \small{startserviceintent} & 43\\
\hline
\small{you have cleared all my queries} & \small{OTHER} & 22\\
\hline
\small{i want view my bill} & \small{viewbillsintent} & 18\\
\hline
\small{i want to phone service} & \small{startserviceintent} & 8\\
\hline
\small{i need help} & \small{OTHER} & 5\\
\hline
\small{i wanna buy a new purchase from you} & \small{startserviceintent} & 5\\
\hline
\small{i purchase plan} & \small{startserviceintent} & 2\\
\hline
\small{i request you to sign up to new connection} & \small{startserviceintent} & 1\\
\hline
\small{i want to know my data usage bills} & \small{viewdatausageintent} & 1\\
\hline
\small{i need intrnet service} & \small{startserviceintent} & 1\\
\hline
\small{my bill keeps going up} & \small{viewbillsintent} & 1\\
\hline
\small{i want cancel cable service} & \small{cancelserviceintent} & 1\\
\hline
\small{internet is very slow} & \small{OTHER} & 1\\
\hline
\small{bill keeps going up} & \small{viewbillsintent} & 1\\
\hline
\small{i liked phone connetion} & \small{startserviceintent} & 1\\
\hline
\end{tabular}
\end{table}

\begin{table}[ht!]
\caption{Manual mapping for \textit{insurance} from top clusters to intents with estimated volumes on test set.}
\label{table:insurance_map}
\centering
\begin{tabular}{|p{2.5cm}|c|c|}
\hline
Top Cluster & Intent & Volume \\
\hline
\small{i want proof of insurance for my car} & \small{getproofofinsurance} & 221\\
\hline
\small{i need status of my claim} & \small{checkclaimstatus} & 114\\
\hline
\small{my phone screen is broken} & \small{reportbrokenphone} & 103\\
\hline
\small{the year is also wrong} & \small{OTHER} & 10\\
\hline
\small{i meet this work} & \small{OTHER} & 8\\
\hline
\small{i need your help} & \small{OTHER} & 4\\
\hline
\small{i need my cliam status} & \small{checkclaimstatus} & 2\\
\hline
\small{my ssn number is lost} & \small{OTHER} & 1\\
\hline
\small{i have bought a new car} & \small{OTHER} & 1\\
\hline
\small{flxing screen} & \small{reportbrokenphone} & 1\\
\hline
\small{my car met accident and bumper was damaged} & \small{OTHER} & 1\\
\hline
\small{please fix my sreen} & \small{reportbrokenphone} & 1\\
\hline
\end{tabular}
\end{table}

\begin{table}[ht!]
\caption{Manual mapping for \textit{finance} from top clusters to intents with estimated volumes on test set.}
\label{table:finance_map}
\centering
\begin{tabular}{|p{2.5cm}|c|c|}
\hline
Top Cluster & Intent & Volume \\
\hline
\small{my credit card was lost} & \small{reportlostcard} & 120\\
\hline
\small{i want to change address on my account} & \small{updateaddress} & 88\\
\hline
\small{need to check my account balance} & \small{checkbalance} & 64\\
\hline
\small{i have error on my credit card bill} & \small{disputecharge} & 60\\
\hline
\small{i need help from you} & \small{OTHER} & 26\\
\hline
\small{i want to transfer money to another account} & \small{transfermoney} & 25\\
\hline
\small{i want to close my account} & \small{closeaccount} & 13\\
\hline
\small{i wanted to inquire whether new lower rates are availble to me} & \small{checkoffereligibility} & 13\\
\hline
\small{my old card expired} & \small{replacecard} & 9\\
\hline
\small{i need to block my credit card} & \small{reportlostcard} & 8\\
\hline
\small{i need transfer money from my a/c to another} & \small{transfermoney} & 8\\
\hline
\small{want to know the banks routing number} & \small{getroutingnumber} & 8\\
\hline
\small{i made a purchase of \$400} & \small{OTHER} & 7\\
\hline
\small{i need to order checks} & \small{orderchecks} & 5\\
\hline
\small{some amount debited my card} & \small{disputecharge} & 4\\
\hline
\small{my cars missed} & \small{reportlostcard} & 4\\
\hline
\small{need to check my balance in my a/c} & \small{checkbalance} & 4\\
\hline
\small{there is an error} & \small{OTHER} & 3\\
\hline
\small{i need a education loan} & \small{OTHER} & 3\\
\hline
\small{i have cheack my balance} & \small{checkbalance} & 2\\
\hline
\small{i have check my account} & \small{OTHER} & 1\\
\hline
\small{i want transfer} & \small{transfermoney} & 1\\
\hline
\small{i need to check my balanace} & \small{checkbalance} & 1\\
\hline
\end{tabular}
\end{table}

\begin{table}[ht!]
\caption{Manual mapping for \textit{software} from top clusters to intents with estimated volumes on test set.}
\label{table:software_map}
\centering
\begin{tabular}{|p{2.5cm}|c|c|}
\hline
Top Cluster & Intent & Volume \\
\hline
\small{i want to reimburse my travel expense} & \small{expensereport} & 119\\
\hline
\small{my skype application is not working} & \small{reportbrokensoftware} & 48\\
\hline
\small{i need to buy keyboards} & \small{startorder} & 47\\
\hline
\small{my outlook software is not working properly} & \small{reportbrokensoftware} & 40\\
\hline
\small{i want check the global status of servers} & \small{checkserverstatus} & 32\\
\hline
\small{i want to set up new recurring orders} & \small{startorder} & 24\\
\hline
\small{please check my puchasing item} & \small{OTHER} & 21\\
\hline
\small{please provide dollar value for food and hotel} & \small{OTHER} & 20\\
\hline
\small{i need software update} & \small{softwareupdate} & 19\\
\hline
\small{my application not working properly} & \small{reportbrokensoftware} & 18\\
\hline
\small{i want buy a musical instruments} & \small{startorder} & 14\\
\hline
\small{i need this model psr-e363} & \small{startorder} & 10\\
\hline
\small{missing periodic software updates} & \small{softwareupdate} & 10\\
\hline
\small{my travel expenses is very high} & \small{OTHER} & 9\\
\hline
\small{i need a help} & \small{OTHER} & 8\\
\hline
\small{i need a some information} & \small{OTHER} & 6\\
\hline
\small{my whatsapp messages are not send} & \small{reportbrokensoftware} & 5\\
\hline
\small{i need travel ticket toreimbursement} & \small{expensereport} & 4\\
\hline
\small{i want reorder basic keybords 10 pieces to 15 pieces} & \small{startorder} & 4\\
\hline
\small{i want report on your software on outlook} & \small{reportbrokensoftware} & 3\\
\hline
\small{what is the procedure to summit my expenses} & \small{expensereport} & 2\\
\hline
\small{i want to cancel one item in my order list} & \small{stoporder} & 2\\
\hline
\small{facing the server error} & \small{checkserverstatus} & 2\\
\hline
\small{i need to report my travel expanses} & \small{expensereport} & 1\\
\hline
\small{i want finance help} & \small{getpromotion} & 1\\
\hline
\small{hike message is not received} & \small{reportbrokensoftware} & 1\\
\hline
\small{i have an issue in my skpe app} & \small{reportbrokensoftware} & 1\\
\hline
\end{tabular}
\end{table}

\begin{figure*}[ht!]
    \centering
    \includegraphics[width=\linewidth]{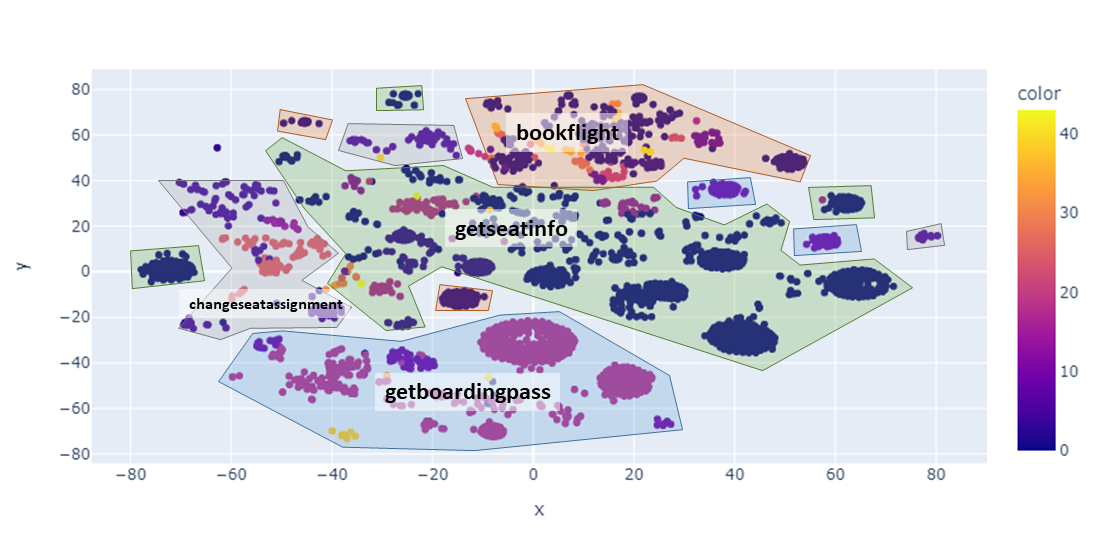}
    \caption{TSNE 2D plot for the \textit{airline} domain with the colorbar as the top-level cluster indicator and the surface area color as the manually defined cluster. The names are intents from the testset with a support higher than 10.}
    \label{fig:airline_clusters}
\end{figure*}

\begin{figure*}[ht!]
    \centering
    \includegraphics[width=\linewidth]{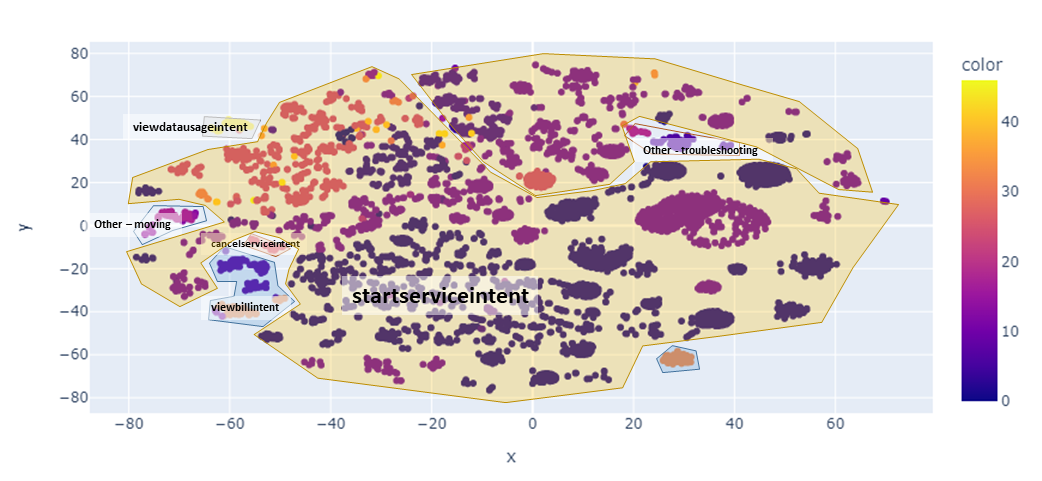}
    \caption{TSNE 2D plot for the \textit{media} domain with the colorbar as the top-level cluster indicator and the surface area color as the manually defined cluster. The names are intents from the testset with a support higher than 10.}
    \label{fig:media_clusters}
\end{figure*}

\begin{figure*}[ht!]
    \centering
    \includegraphics[width=\linewidth]{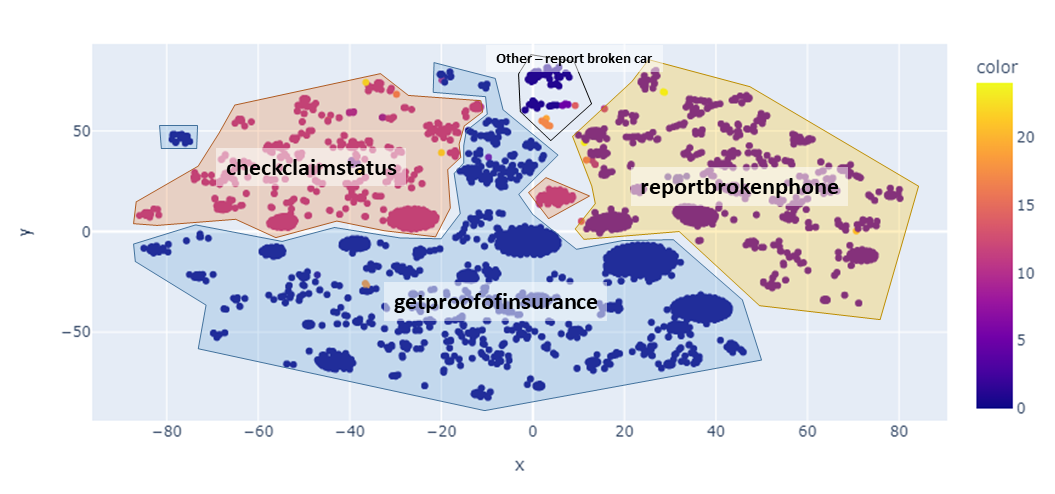}
    \caption{TSNE 2D plot for the \textit{insurance} domain with the colorbar as the top-level cluster indicator and the surface area color as the manually defined cluster. The names are intents from the testset with a support higher than 10.}
    \label{fig:insurance_clusters}
\end{figure*}

\begin{figure*}[ht!]
    \centering
    \includegraphics[width=\linewidth]{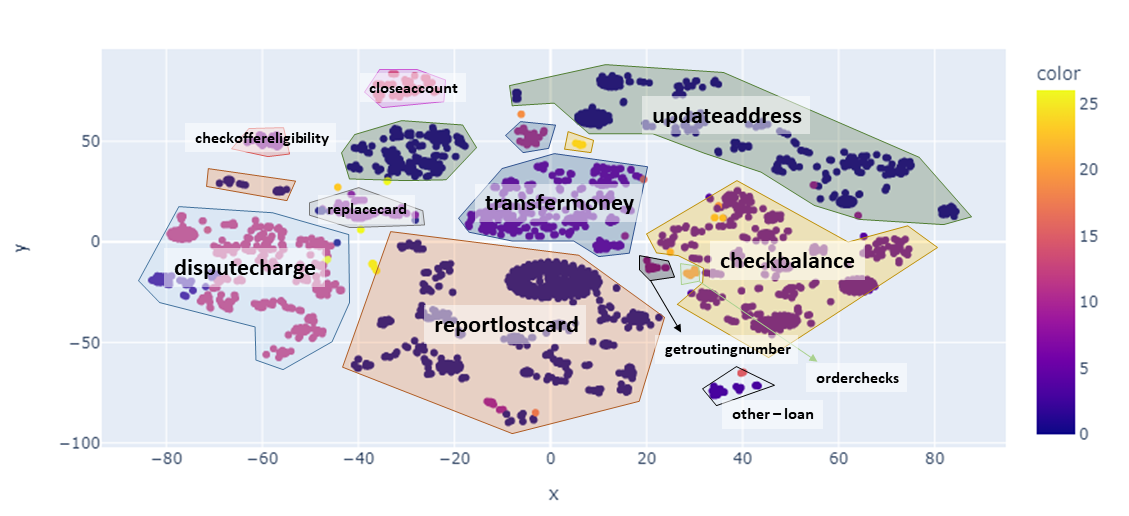}
    \caption{TSNE 2D plot for the \textit{finance} domain with the colorbar as the top-level cluster indicator and the surface area color as the manually defined cluster. The names are intents from the testset with a support higher than 10.}
    \label{fig:finance_clusters}
\end{figure*}

\begin{figure*}[ht!]
    \centering
    \includegraphics[width=\linewidth]{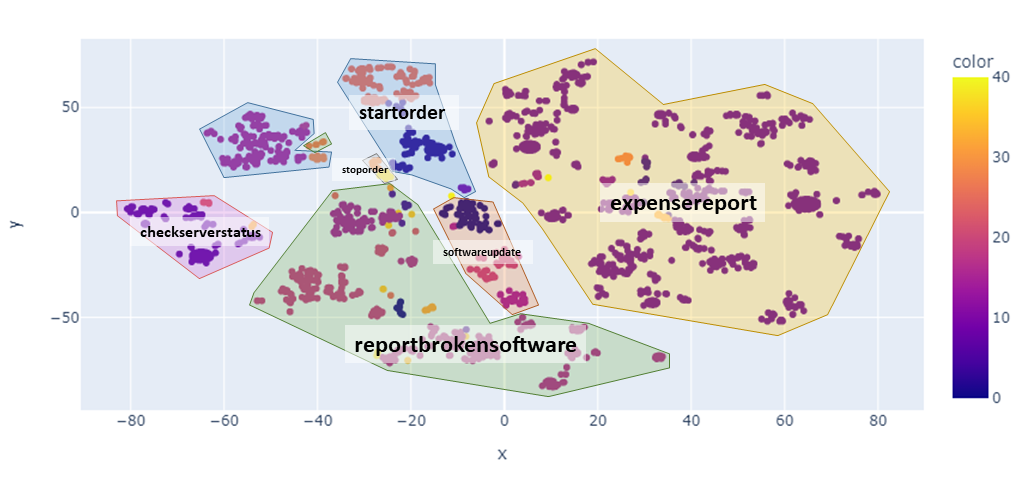}
    \caption{TSNE 2D plot for the \textit{software} domain with the colorbar as the top-level cluster indicator and the surface area color as the manually defined cluster. The names are intents from the testset with a support higher than 10.}
    \label{fig:software_clusters}
\end{figure*}

\begin{table}[ht!]
    \caption{Cluster hyperparameters.}
    \centering
    \begin{tabular}{|c|c|c|c|}
        \hline
         & \small{\textit{min\_cluster}} & \small{\textit{distance}} & \small{\textit{force\_cluster}}\\
         & \small{\textit{\_size}} & \small{\textit{\_threshold}} & \small{\textit{\_threshold}}\\
        \hline
        \small{airline} & 4 & 0.29 & 0.3\\
        \hline
        \small{media} & 3 & 0.42 & 0.2\\
        \hline
        \small{insurance} & 2 & 0.5 & 0.2\\
        \hline
        \small{finance} & 2 & 0.45 & 0.2\\
        \hline
        \small{software} & 2 & 0.5 & 0.3\\
        \hline
    \end{tabular}
\label{tab:cluster_hp}
\end{table}

\end{document}